\documentclass[conference]{IEEEtran}

\IEEEoverridecommandlockouts
\usepackage{cite}
\usepackage{amsmath,amssymb,amsfonts}
\usepackage{algorithmic}
\usepackage{graphicx}
\usepackage{textcomp}
\usepackage{soul}
\usepackage[table,xcdraw]{xcolor}
\usepackage{hyperref}
\def\BibTeX{{\rm B\kern-.05em{\sc i\kern-.025em b}\kern-.08em
    T\kern-.1667em\lower.7ex\hbox{E}\kern-.125emX}}

\makeatletter
\def\ps@IEEEtitlepagestyle{%
	\def\@oddfoot{\mycopyrightnotice}%
	\def\@evenfoot{}%
}
\def\mycopyrightnotice{%
	{\hfill \footnotesize 978-1-6654-5705-7/23/\$31.00~\copyright 2023 IEEE\hfill}
}
\makeatother

\begin{document}

\title{Federated Learning for Early Dropout Prediction on Healthy Ageing Applications

\thanks{This publication has been partially funded by the European Union's Horizon 2020 research and innovation programme under the TERMINET project with Grant Agreement No. 957406.}
}

\makeatletter
\newcommand{\linebreakand}{%
  \end{@IEEEauthorhalign}
  \hfill\mbox{}\par
  \mbox{}\hfill\begin{@IEEEauthorhalign}
}
\makeatother

\author{
    \IEEEauthorblockN{
       Christos Chrysanthos Nikolaidis\IEEEauthorrefmark{1},
       Vasileios Perifanis\IEEEauthorrefmark{1}, 
       Nikolaos Pavlidis\IEEEauthorrefmark{1} and
       Pavlos S. Efraimidis\IEEEauthorrefmark{1}, 
    }
    \IEEEauthorblockA{
        \IEEEauthorrefmark{1} Department of Electrical and Computer Engineering, Democritus University of Thrace, Xanthi, Greece \\
        }
    \IEEEauthorblockA{e-mail: \{cnikolai, vperifan, npavlidi, pefraimi\}@ee.duth.gr}
    
}


\newcommand{\vasilis}[1]{\textcolor{red}{ #1}}
\newcommand{\nikos}[2]{\textcolor{blue}{ #1}}
\newcommand{\chris}[1]{\textcolor{cyan}{#1}}

\maketitle
\thispagestyle{plain}
\pagestyle{plain}

\begin{abstract}

The provision of social care applications is crucial for elderly people to improve their quality of life and enables operators to provide early interventions. Accurate predictions of user dropouts in healthy ageing applications are essential since they are directly related to individual health statuses. Machine Learning (ML) algorithms have enabled highly accurate predictions, outperforming traditional statistical methods that struggle to cope with individual patterns. However, ML requires a substantial amount of data for training, which is challenging due to the presence of personal identifiable information (PII) and the fragmentation posed by regulations. In this paper, we present a federated machine learning (FML) approach that minimizes privacy concerns and enables distributed training, without transferring individual data. We employ collaborative training by considering individuals and organizations under FML, which models both cross-device and cross-silo learning scenarios. Our approach is evaluated on a real-world dataset with non-independent and identically distributed (non-iid) data among clients, class imbalance and label ambiguity. Our results show that data selection and class imbalance handling techniques significantly improve the predictive accuracy of models trained under FML, demonstrating comparable or superior predictive performance than traditional ML models.
\end{abstract}


\section{Introduction}
Digitalization has brought a plethora of possibilities for health and social care operations, including early health interventions \cite{colizzi_prevention_2020}, which support individuals by enabling self-care practices \cite{TRIANTAFYLLIDIS2015743}. One of the most impactful applications of this digital transformation is the early identification of patterns that indicate an individual might dropout (discontinue) a treatment program. Such applications can significantly enhance individuals' quality of life, e.g., by minimizing the risk of disease development \cite{wiens_no_2019} and identifying signs of depression \cite{burns_harnessing_2011}. 

Building on these advances, the implementation of smart health and social care applications empowers individuals, especially the elderly people, to enhance their overall well-being \cite{QIN2020104121}. Such digital tools are particularly crucial for elderly people, as they monitor their physical and cognitive activities, having a vital role in improving their health status.

By integrating machine learning (ML) techniques into these applications, intelligent services can be developed, resulting in substantial advantages for individuals and reduced economic costs for health operators. However, ML algorithms require large amounts of training data to provide highly accurate predictions. The transmission of data that, in the context of health and social care, contain personally identifiable information (PII) to a central server (CS) as well as data sharing among organizations, face limitations due to the increased privacy concerns and the fragmentation posed by regulations such as the GDPR \cite{Voigt_2017}. Hence, research efforts have recently turned towards the decentralization of ML, with federated machine learning (FML) emerging as an attractive research area that enables model training without data transmission. 

FML, introduced by McMahan et al. \cite{pmlr-v54-mcmahan17a}, allows the training data to remain distributed on edge devices while a shared model is learned by securely aggregating client-based computed updates \cite{10.1145/3133956.3133982}. Unlike traditional ML, in FML, a CS is responsible for coordinating participating clients and collecting model updates instead of raw data. This distributed approach ensures that the data remains on the clients' devices, minimizing privacy and security concerns.

Although FML has received much attention in recent years, the final model’s predictive accuracy is affected by several challenges introduced by the distributed data, with the most common problem being data heterogeneity \cite{ZHU2021371, 9835537, perifanis2022federated}. Since FML clients are likely to own different data distributions, careful selection of clients per federated round is crucial for a federated model to converge towards the global objective \cite{pmlr-v151-jee-cho22a}. However, determining which clients should be included or excluded during specific federated training rounds is a complex task that involves addressing various issues and considerations such as resource allocation, data quality and quantity.

Motivated by the above, we propose the application of a FML-based training to predict early dropouts from active and healthy ageing applications. Timely accurate predictions of dropouts enable early interventions by health and social care operators, reduced economical costs associated with hospitalization and ultimately, improve the quality of life of patients, while preventing them from disease development or mood changes. In this paper, we focus on elderly people and train a deep neural network using FML to minimize privacy related issues. 

The main contributions of this paper are as follows:
\begin{itemize}
    \item We explore the effectiveness of FML for predicting early dropouts from monitored interactions within a healthy ageing mobile application, a novel and very promising approach that enables early interventions for improving the overall well-being of individuals.
    \item We comprehensively evaluate the impact of data heterogeneity and class imbalance on the predictive performance of FML.
    \item We compare several data selection and class imbalance handling techniques. 
\end{itemize}

Our findings indicate that by employing data selection and resampling techniques, FML approaches can achieve comparable or even superior performance compared to their centralized counterparts.

The rest of this paper is structured as follows: Section \ref{related} presents the related work. Section \ref{methodology} introduces the dataset, defines the problem and FML architecture and illustrates the use-case scenarios. Section \ref{results} presents the experimental details and compares the FML to centralized learning on different scenarios. Finally, Section \ref{conclusion} summarizes the key outcomes.

\section{Related Work}
\label{related}
\subsection{Machine Learning for Healthcare Applications}
In the current literature, there has been an explosive growth in research efforts that employ ML algorithms to predict individuals’ health status. Due to the absence of social care-based datasets, we present related studies on the application of ML techniques to the broader healthcare domain.

In \cite{QIN2020104121}, the authors predicted individuals' health conditions using self-reports, utilizing ML algorithms such as Support Vector Machines (SVM) and Multi-Layer Perceptron (MLP). Prati et al. \cite{PRATI2022104791} used an interview-based dataset to predict the participants' quality of life, life satisfaction and happiness, employing the Random Forest and Gradient Boosting algorithms. In \cite{torous_smartphones_2018}, the authors targeted early interventions by predicting suicide risks. Finally, O’Keeffe et al. \cite{okeeffe_predicting_2018} focused on predicting the dropouts of individuals receiving therapy for depression to enable early interventions using the Logistic Regression algorithm. Note that all of these studies emphasized the advantages of ML algorithms compared to traditional statistical methods.

Despite ongoing research efforts, the prevailing literature concerning dropout and treatment engagement prediction in this domain predominantly relies on datasets derived from interviews or self-reports. However, such sources lack real-time dimensionality, which is critical for timely accurate assessment of an individual's health condition. To address this issue, Fico et al. \cite{Fico_2023_MAHA} introduced a novel dataset derived from a mobile application designed for elderly people. The application prompts users to engage in various activities within a gamified environment and collects user acquisitions. The long-term goal is to evaluate user engagement through intelligent predictions, enabling social care operators to implement early interventions.

\subsection{Federated Learning for Healthcare Applications}
The application of FML in healthcare applications has great potential to improve privacy and offer social good \cite{rieke_future_2020}. Such use cases has been explored in many research efforts, including \cite{BRISIMI201859, dayan_federated_2021, adnan_federated_2022,  9900733}. More specifically, \cite{BRISIMI201859} used FML to predict heart-related hospitalization. Brisimi et al. \cite{dayan_federated_2021} employed FML across 20 health institutes to predict future oxygen requirements for patients with COVID-19. In \cite{adnan_federated_2022}, the authors trained a deep learning model using FML for medical image analysis. Finally, Diamantoulaki et al. \cite{9900733} trained a fully connected neural network with FML for disease risk assessment at an early stage.

While there are several applications of FML in various use cases within the healthcare domain, to the best of our knowledge, none of these studies specifically address patient engagement or application dropout prediction. We leverage FML to showcase the potential in this unique domain, emphasizing the need for user selection and data imbalance handling techniques, in a real-world scenario.

\section{Methodology}
\label{methodology}

In this section we introduce the real-world social care dataset, we formulate the problem and describe how federated training is performed. Finally, we analyze data selection and resampling techniques employed to address class imbalance issues.

\subsection{Dataset}\label{dataset}
The experiments were conducted on the MAHA dataset \cite{Fico_2023_MAHA},\footnote{The dataset was introduced as a part of the \href{https://wchallenge2022.lst.tfo.upm.es/index.html}{IFMBE Challenge 2022}.} a real-world social care dataset collected between 2018 and 2021. The dataset was generated by conducting an experiment on elderly people in Spain, who were asked to perform various activities in an application. Among these individuals, there were people with disabilities, e.g., kinetic and memory problems. The dataset consists of user acquisitions related to the requested activities, which belong to one of the following categories: Brain Games, Fingertapping, Mindfulness and Physical Activity. The objective of the MAHA application is to support elderly people by requesting them to participate in a gamified environment by performing at least two activities within a one-week interval. By monitoring the individual acquisitions and predicting user adherence, the application aims to improve the quality of life of elderly people and ultimately, to offer early interventions when a participant's engagement with the requested activities is low.

For each individual, the acquisitions are divided into two sessions, Monday to Thursday and Friday to Sunday, as described in
\cite{Perifanis_2023_Dropout}. In addition, a window of twelve past sessions (or six weeks), containing aggregated statistics regarding the user activity is used as input and the goal is to predict whether a user will interact with the application in the subsequent three sessions. If a user performs acquisitions two or more times during these three sessions, a label of one is assigned, indicating high adherence. Otherwise, a label of zero is assigned, suggesting that the individual is likely to dropout of the application. Following \cite{Perifanis_2023_Dropout}, there is a class imbalance, where low adherence corresponds to 75\% of all instances. Finally, the dataset contains label ambiguity, i.e., there are identical feature vectors $x$ that correspond to both low and high adherence labels.

Fig. \ref{fig:user_quantities} illustrates the number of samples per user identifier belonging to the training set along with their corresponding labels. From this figure, it is demonstrated that the observations among users vary significantly in terms of the number of samples and that each individual holds unique patterns. That is, the distribution varies significantly among users. For instance, some users hold a much larger fraction of samples belonging to class 0, while others hold many observations belonging to class 1.

The characteristics of the MAHA dataset, i.e., the inclusion of \textit{sensitive information} of elderly people, the presence of \textit{data diversity} regarding both the acquisition patterns and the demographic variety as well as the presence of \textit{label imbalance} and \textit{label ambiguity} make it an appealing real-world dataset for experimenting with FML. Our primary objective is to assess whether FML can achieve highly accurate predictions in real-world scenarios involving the aforementioned heterogeneous characteristics.

\begin{figure}
    \centering
    \includegraphics[width=\columnwidth]{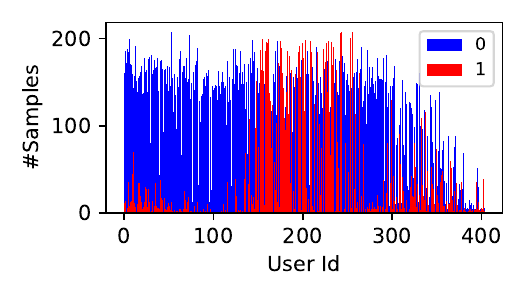}
    \caption{Number of samples per label and user on the MAHA dataset.}
    \label{fig:user_quantities}
\end{figure}

\subsection{Problem Definition and FML application}
We consider a network of $N$ FML participating clients and a central server, responsible for coordinating the training process and aggregating model parameters from the devices. Note that each client $n \in N$ can be either an individual user or a social care operator responsible for collecting data from their monitoring individuals. Each client owns a local dataset that consists of user acquisitions $\mathcal{D}_n = \bigcup_{i \in \mathcal{O}_m}\{x_i, y_i\}$, where $x_i$ is a vector containing the number of acquisitions per session within a window $T$, $y_i$ is the the corresponding binary label and $\mathcal{O}_m$ denotes the observations for a client monitoring $m$ users. The entire dataset consisting of all observations is denoted as $\mathcal{D} = \bigcup_{j \in n} \mathcal{D}_j$. Note that each $x_i$ comprises $T$ entries and in the considered use case, we set $T=12$. In addition, $y_i$ is a binary label indicating low or high adherence and is generated based on the future observations $x_{i+1:i+p, T}$. Specifically, we set $p=3$ and $y_i = 0$ if $\sum_{j=1}^{p} x_{i+j, T} < 2$ and $y_i = 1$, otherwise. The goal for the $N$ clients is to come up with a common predictor, defined as:
\begin{equation} \label{eq:pred}
    f(\theta, x_i^n) = (y_i^n)',  \forall n \in N,
\end{equation}
where $f(\theta, x_i^n)$ is a learning model with parameters $\theta$, $x_i^n$ is the $i$th input vector from the $n$th client and $(y_i^n)'$ is the corresponding prediction. During training, each client $n \in N$ minimizes a local loss function:
\begin{equation} \label{eq:loss}
    F(\theta^n) = \frac{1}{\mathcal{D}_n} \sum_{i=1}^{|\mathcal{D}_n|} (f(\theta, x_i^n, y_i^n)),
\end{equation}
where $f(\theta, x_i^n, y_i^n)$ captures the error between the ground truth $y_i^n$ and the predicted value $(y_i^n)'$. Each device updates the local model parameters $\theta$ to minimize $F(\theta^n)$ using stohastic gradient descent (SGD):
\begin{equation} \label{eq:sgd}
    \theta_{t+1}^n = \theta_t^n - \gamma \nabla F(\theta_t^n),
\end{equation}
where $\theta_t^n$ is the calculated parameters at iteration $t$ on the $n$th device, $\gamma$ is the learning rate and $\nabla F(\theta_t)^n$ is the gradient of the loss with respect to the model parameters. The training process should capture a global optimization objective and generalize among all $N$ clients, i.e.,
\begin{equation} \label{eq:global_obj}
    F(\theta) = \frac{1}{\mathcal{D}} \sum_{n=1}^{N} \mathcal{D}_n F(\theta).
\end{equation}

To generate a common predictor, we employ a federated learning architecture as shown in Fig. \ref{fig:fl-architecture}. In the first federated round, the CS initializes the weights and biases of a neural network, denoted as $\theta_0$. During federated training the following steps are repeated for an arbitrary round $e$ \cite{pmlr-v54-mcmahan17a, 10.1145/3133956.3133982}:
\begin{itemize}
    \item Step 1: The CS samples available clients $n_e \in N$ based on a selection function $S(c)$, where $c$ is a selection fraction. In this work, we use a random sampling approach. After client selection, the CS transmits the current global model parameters $\theta_e$ to the selected set of devices $n_e$.
    \item Step 2: The selected clients $k \in n_e$ receive the current parameters $\theta_e$ and apply local training iterations using Eq.~\ref{eq:sgd}, optimizing a loss function (Eq. \ref{eq:loss}). After local training, they transmit the updated model parameters $\theta_{e+1}^k$ to the CS.
    \item Step 3: After receiving the local parameters from each $k \in n_e$, the CS securely aggregates the parameters and forms the new global parameters $\theta_{e+1}$. In this work, we use the FedAvg's aggregation \cite{pmlr-v54-mcmahan17a}:
    \begin{equation}
        \theta_{e+1} = \dfrac{1}{\sum_{k \in n_e} |D_k|} \sum_{k \in n_e} |D_k| \theta_{e+1}^k
    \end{equation}
\end{itemize}

The above process repeats until the local errors (Eq. \ref{eq:loss}) minimize as much as possible without global model overfitting, i.e., the predictor can generalize among $N$ clients (Eq. \ref{eq:global_obj}) and provide highly accurate predictions (Eq. \ref{eq:pred}).

\begin{figure}[t!]
\centerline{\includegraphics[width=0.95\columnwidth]{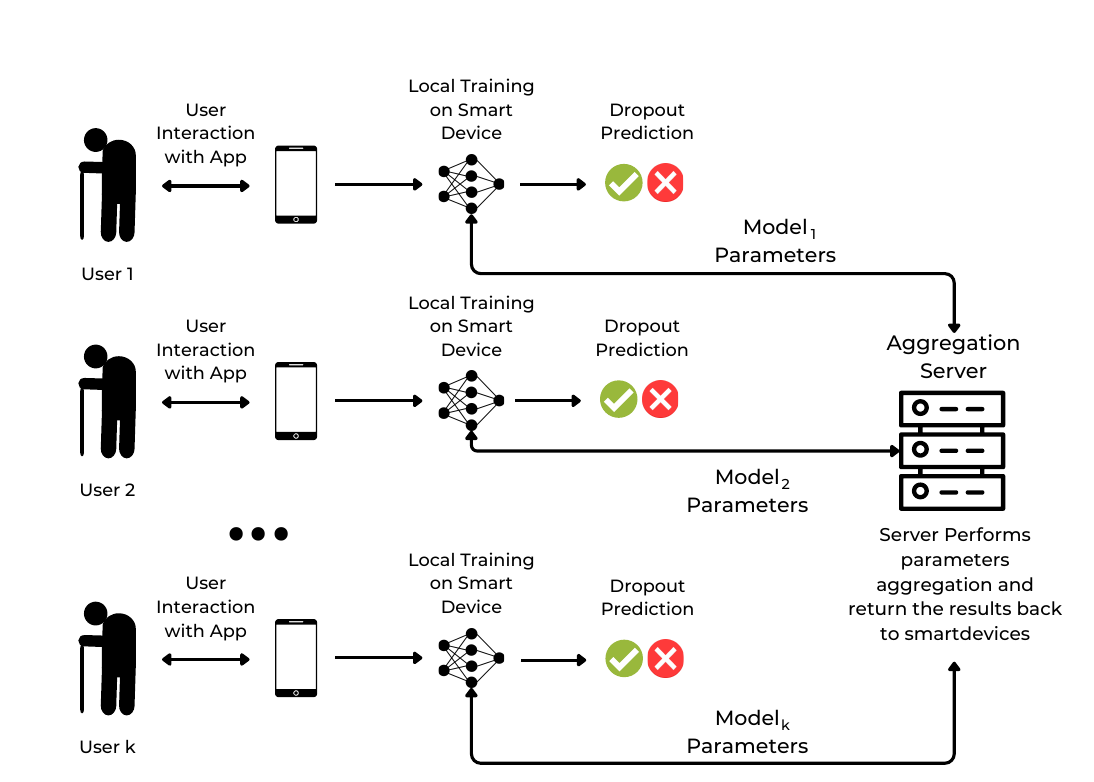}}
\caption{FL with Individual Participants - Architecture}
\label{fig:fl-architecture}
\end{figure}

\subsection{Data Selection and Class Imbalance Handling}
Selecting appropriate data, or users in the context of FML, and handling class imbalance are critical in generating high accurate ML models \cite{CAI201870}. Under FML, where data distributions can vary significantly among devices, selecting appropriate data can help in solving non-iid related problems \cite{ZHU2021371}. To assess the impact of data quality and class imbalance in the considered use case, we consider the following experimental scenarios:
\begin{itemize}
    \item \textbf{Raw Data}: In this experiment, we use the original dataset without any data selection and imbalance handling techniques. The goal of this scenario is to understand the characteristics and patterns in the dataset without any modifications.
    \item \textbf{User Exclusion based on Low Information Content}: In this experiment, we filter out clients with a limited number of observations for either low (0) or high (1) adherence labels. Such sparse data can be less informative and introduce noise during training deep learning algorithms. More precisely, we consider two thresholds: 5 and 10. Clients with fewer than these numbers of samples for labels `0' or `1' are excluded from training. This filtering ensures that only users with a significant representation of both labels are included in the training stage.   
    \item \textbf{Over Sampling}: In this experiment, we handle class imbalance by increasing the representation of the minority class (high adherence) using the k-Means Synthetic Minority Oversampling Technique (k-MeansSmote) \cite{DOUZAS20181}.
    \item \textbf{Under Sampling}: In this experiment, we handle class imbalance by reducing the representation of the majority class (low adherence). We employed a random under sampling method for this purpose.
    \item \textbf{Over-Under Sampling}: In this scenario, we handle class imbalance by considering a combination of over sampling and under sampling. We augment the minority class by generating synthetic samples via over sampling. We also reduce the samples in the majority class through under sampling to achieve balance. We utilize the Synthetic Minority Oversampling Technique and Edited Nearest Neighbor (SMOTE-ENN) algorithm \cite{10.1145/1007730.1007735} for this approach.
\end{itemize}


\section{Experiments}
\label{results}
In this section, we experimentally compare three learning settings, i.e., centralized, cross-device FML and cross-silo FML, to demonstrate the effectiveness of ML algorithms in predicting early user dropouts. We outline the experimental setup and the three learning settings and we present the experimental results.

\subsection{Learning Settings}
In our experiments, we evaluated three distinct scenarios: i) centralized learning, ii) cross-device FML, where each client represents a single individual and iii) cross-silo FML, where each client holds observations from multiple individuals, e.g., a hospital or a social care operator. More specifically, we consider the following learning settings:
\begin{itemize}
    \item \textbf{Centralized Learning}: In traditional ML, the end-users transmit their data to a CS, where the learning process occurs. In this setting, user privacy is not maintained, as users' data becomes accessible to a central entity. Nonetheless, centralized learning often exhibits superior predictive accuracy to distributed learning settings \cite{9306745} and serves as the upper bound for FML.
    \item \textbf{Cross-Device Federated Learning}: Given the highly sensitive nature of the personal and confidential data involved in the application, we employ FML as a sophisticated approach to enhance privacy and ensure secure data handling. In cross-device FML, we treat each end-user as an independent client, ensuring that their local data remains solely on the local device. Each user conducts local training using their private data and subsequently transmits the local model weight's to the CS.
    \item \textbf{Cross-Silo Federated Learning}: We divided the individual participants into three groups by randomly selecting users from the dataset. This approach resembles the scenario in which end-user data are monitored and collected by another entity like a social care operator. The first and second group consists of 134 individuals, respectively, while the third group comprises 136 clients.
\end{itemize}

For all learning settings, we consider a fully connected neural network. Specifically, the input layer is followed by three Dense layers with 128, 64 and 32 units, respectively. The activation function among hidden layers is rectified linear unit (ReLU). In addition, a dropout \cite{srivastava2014dropout} of 0.2 is applied after each dense layer to regularize the model and reduce overfitting. Finally, the prediction is generated by a Dense layer with a single unit using the sigmoid activation function.

The dataset described in Section \ref{dataset} comprises a total of 454 individuals. For all of our experiments, we excluded 50 individuals from training at random and used their observations for testing. The remaining individuals were then used for training the models. This process ensures that the selected individuals are held out and have never been seen during training. To obtain generalized and robust results, we trained the model 10 times from scratch using different seed values. 

For the two FML scenarios, we utilized the FLOWER \cite{beutel2020flower} framework as the underlying infrastructure and Tensorflow \cite{tensorflow2015-whitepaper} as the backbone deep learning library. We conducted a total of 20 federated rounds to train and update the global model parameters. During each federated round, the participating clients perform local training by conducting 5 local epochs using their local datasets. In the cross-device FML setting, at each federated round the CS samples the 40\% ($c=0.4$) of total users, while for the cross-silo FML, the CS selects all clients (in this case, three). For the centralized setting, we set a total of 20 epochs and for all settings we use a learning rate of $10^{-3}$ with a batch size of 32, optimizing the binary cross entropy (BCE) loss.
    

\subsection{Metrics}
We employed several performance metrics to evaluate the effectiveness of the ML models. These metrics include Accuracy (Acc), Precision (P), Recall (R), F1 score, and Geometric Mean (GM). Accuracy represents the proportion of correct predictions, while Precision measures the proportion of positive predictions that were actually positive. Recall quantifies the proportion of positive instances that were correctly predicted as positive. F1 score combines Precision and Recall into a single metric using their harmonic mean. Finally, GM is the geometric mean of Sensitivity (or Recall) and Specificity, which measures the ability of a classifier to correctly predict negative cases. The considered metrics are defined as follows:
\begin{equation}
    Acc = \dfrac{TP + TN}{TP + FP + TN + FN}
\end{equation}
\begin{equation}
    P = \dfrac{TP}{TP + FP}
\end{equation}
\begin{equation}
    R = \dfrac{TP}{TP + FN}
\end{equation}
\begin{equation}
    F1 = \dfrac{2 \times P \times R}{P + R}
\end{equation}
\begin{equation}
    GM = \sqrt{\dfrac{TP}{TP + FN} \times \dfrac{TN}{TN + FP}},
\end{equation}
where $TP$ and $TN$ are the actual number of positive (1) and negative (0) samples that are correctly predicted as positive and negative, respectively; $FP$ is the number of negative samples that incorrectly predicted as positive and $FN$ is the number of positive cases that incorrectly predicted as negative.

\begin{table*}[htbp]
\caption{Results Across Learning Setting On Test Sets.}
\centering
\begin{tabular}{l|ccccc|ccccc|ccccc}
\hline
                       & \multicolumn{5}{c|}{\textbf{Centralized}} & \multicolumn{5}{c|}{\textbf{Cross-Device Federated Learning}} & \multicolumn{5}{c}{\textbf{Cross-Silo Federated Learning}} \\ \hline
Experiment              & Acc    & P     & R    & F1        & GM        & Acc      & P     & R    & F1        & GM      & Acc     & P   & R    & F1        & GM              \\ \hline        

Raw data                & 0.86        & 0.76          & 0.67      & 0.71      & 0.79      & 0.82          & 0.92          & 0.32      & 0.47      & 0.56    & 0.78          & 0.94        & 0.15      & 0.26      & 0.38            \\ \hline

Drop Users (5)          & 0.86        & 0.77          & 0.66      & 0.71      & 0.79      & 0.86          & 0.78          & 0.63      & 0.69      & 0.77    & 0.83          & 0.77        & 0.48      & 0.59      & 0.68             \\ \hline 
Drop Users (10)         & 0.86        & 0.76          & 0.68      & 0.72      & 0.79      & 0.85          & 0.74          & 0.68      & 0.71      & 0.79    & 0.79          & 0.75        & 0.29      & 0.41      & 0.52             \\ \hline
Over sampling            & 0.79        & 0.57          & 0.68      & 0.62      & 0.75      & 0.81          & 0.81          & 0.36      & 0.49      & 0.58    & 0.68          & 0.23        & 0.09      & 0.13     & 0.29             \\ \hline 
Under sampling           & 0.80        & 0.60          & 0.73      & 0.66      & 0.78      & 0.84          & 0.68          & 0.76      & 0.71      & 0.81    & 0.72          & 0.47        & 0.77      & 0.58      & 0.73             \\ \hline
Over \& Under Sampling  & 0.81        & 0.65          & 0.59      & 0.62      & 0.72      & 0.85          & 0.75          & 0.65      & 0.69      & 0.77    & 0.71          & 0.38        & 0.18      & 0.25      & 0.40             \\ \hline
\end{tabular}
\label{tab:results}
\end{table*}

\subsection{Results}
\begin{figure}[t!]
\centering{\includegraphics[width=\columnwidth]{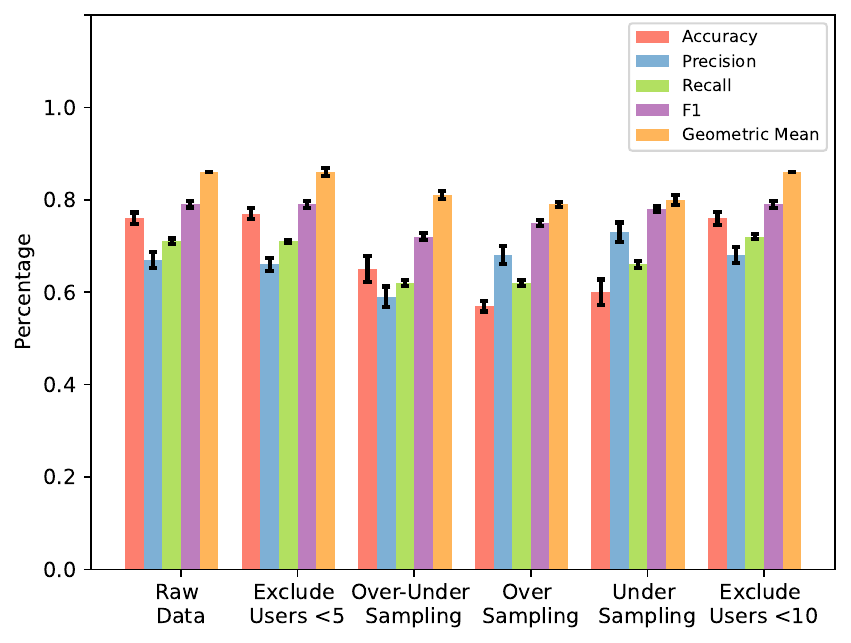}}
\caption{Results from centralized experiments per data selection scenario.}
\label{fig:results_cen}
\end{figure}
The results for all metrics and learning settings are presented in Table \ref{tab:results}. The following observations are made per learning setting:
\addtolength{\leftmargini}{1cm}
\begin{itemize}
    \item \textbf{Centralized Learning}: 
    When applied to raw data, the centralized approach achieved an Accuracy of 86\%, Precision of 76\%, Recall of 67\%, F1 score of 71\% and a Geometric Mean of 79\%. With the utilization of re-sampling techniques to address the influence of class imbalance such as over sampling, undersampling and over-under sampling, we observed a decline in the predictive accuracy. This decrease is attributed to the presence of label ambiguity in our dataset (Section~\ref{dataset}), where the same features can be associated with both low and high adherence labels. In addition, when we excluded users with low informative content, there was not a significant shift in the model's predictive performance. The stability of the results across different settings is linked to the centralized model's capability to access all user data during training. The average metrics and their standard deviation for each technique under centralized learning are summarized in Fig. \ref{fig:results_cen}.
    
    \item \textbf{Cross-Device Federated Learning:} 
    In cross-device FML, the results varied from those observed in the centralized learning setting. Specifically, on the raw data, the model achieved an Accuracy of 82\%, Precision of 92\%, Recall of 32\%, F1 score of 47\% and a Geometric Mean of 56\%. These results demonstrate a reduction of 5\% in Accuracy, 52\% in Recall, 34\% in F1 score, and 29\% in the Geometric Mean when compared to the centralized model. This decrease is attributed to data heterogeneity across individual participants.    
    
    
    Different from the centralized setting, the data selection and class imbalance handling techniques showed mixed results. More precisely, the application of over-under sampling in cross-device FML demonstrated a slight improvement of 10\% in Recall, 11\% in F1 Score and 7\% in Geometric Mean compared to the centralized counterpart. Similarly, the under sampling approach resulted in higher accuracy results when compared to the corresponding centralized experiment. In contrast, the introduction of oversampling led to a notable reduction in metrics. Specifically, there is a decline of 47\% in Recall, 21\% in F1 Score, and 23\% in Geometric Mean compared to the outcomes of the centralized experiment. This behavior can be attributed to the high quantity skew of data samples, i.e., varying number of samples, which is present among clients. Characteristically, 85 users own under 100 observations, while a mere 79 has over 200 instances. This differentiation makes it challenging for over sampling algorithms to generate accurate synthetic data and consequently improve the overall predictive performance of the deep learning algorithm. 
    
    After thoroughly analyzing the results in the cross-device context and contrasting them with the raw data scenario, the findings diverged from initial expectations. Specifically, the utilization of under-sampling increased the Recall by 138\%, the F1 score by 51\%, and the GM by 45\%. Notably, when the cross-device FML is combined with the under-sampling method, it outperforms all other learning scenarios potentially by better capturing irregular patterns among the data distribution.

           
    
    \begin{figure}[t!]
    \centering{\includegraphics[width=\columnwidth]{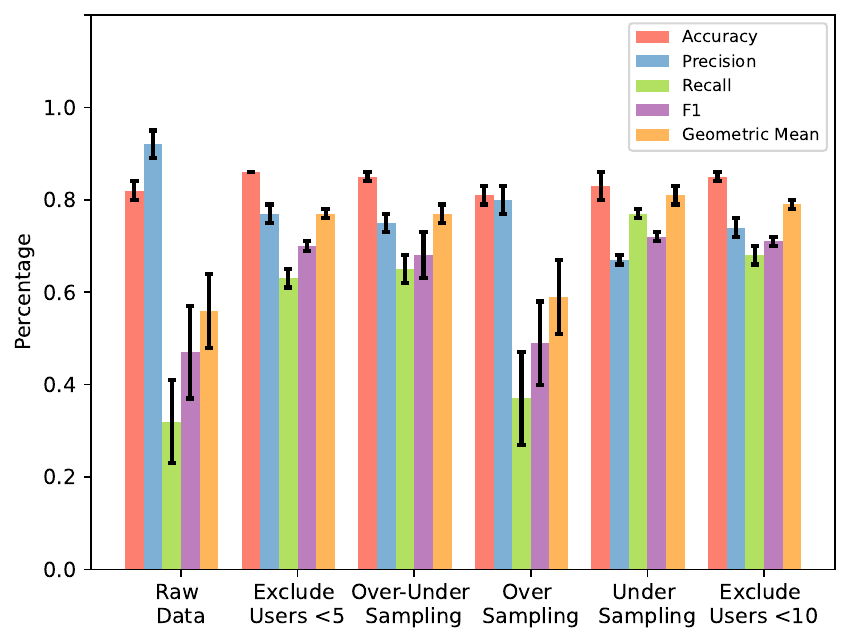}}
    \caption{Results from FL with individuals participants per data selection scenario.}
    \label{fig:results_fl_ind}
    \end{figure}
    Using the data selection approaches, i.e., by excluding users with limited observations for both labels, leads to an improvement in the Recall metric by 97\%, F1 score by 47\%, and Geometric Mean by 38\% compared to raw data training. This suggests that FML models are more susceptible to low quality or noisy content than centralized learning. Since FML models can only access data subsets, the presence of noisy data leads to decreased predictive performance.
    
    
    Fig. \ref{fig:results_fl_ind} summarizes the obtained average metrics along with their standard deviation per considered technique in the cross-device FML setting.
    
    \begin{figure}[b!]
    \centerline{\includegraphics[width=\columnwidth]{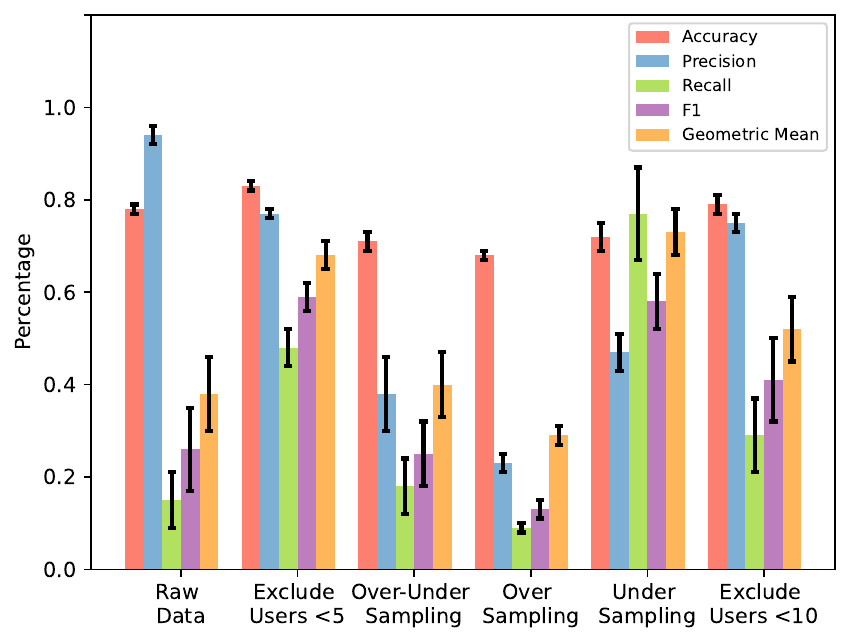}}
    \caption{Results from FL with groups of participants per data selection scenario.}
    \label{fig:results_fl_groups}
    \end{figure}
    \item \textbf{Cross-Silo Federated Learning:}
    As shown in Table \ref{tab:results}, the cross-silo setting yielded the least predictive accuracy. In particular, using the raw data, the model achieves an Accuracy of 78\%, Precision of 94\%, Recall of 15\%, F1 score of 26\% and Geometric Mean of 38\%. These results indicate that the model was not able to correctly identify a majority of samples belonging to the high adherence class. However, when the model did successfully predict instances in this class, its predictions were mostly accurate.
    
   When we exclude individuals with low informative content based on the threshold of 5, the model notably improves the considered metrics. More precisely, the Recall value increased by 220\%, the F1 Score by 127\%, and the Geometric Mean by 79\%.  On the other hand, when we exclude users using the threshold of 10, the predictive performance was decreased in comparison to exluding based on the threshold of 5. This finding suggest that using a threshold for excluding low-informative clients can improve the overall performance of the FML model. However, this threshold is specific to the implementation and requires an exploratory analysis to determine its optimal value.
   
    
    Regarding re-sampling techniques, we found that overs sampling had a significant negative impact on the performance metrics, decreasing Recall by 40\%, F1 score by 50\%, and Geometric Mean by 24\%. On the other hand, under sampling had a positive impact, increasing Recall by 413\%, F1 score by 123\% and Geometric Mean by 92\%. Finally, over-under sampling had an improve in Recall by 20\%, but a slightly decrease in Accuracy by 9\%. Fig. \ref{fig:results_fl_groups} illustrates the average metrics along with their standard deviation per considered technique in the cross-silo FML setting.
    

    Despite the significant improvements achieved through the use of under sampling or data selection techniques in cross-silo FML, the predictive accuracy still falls short compared to the centralized and cross-device FML settings. 
    
\end{itemize}
\begin{figure}[t!]
\centerline{\includegraphics[width=\columnwidth]{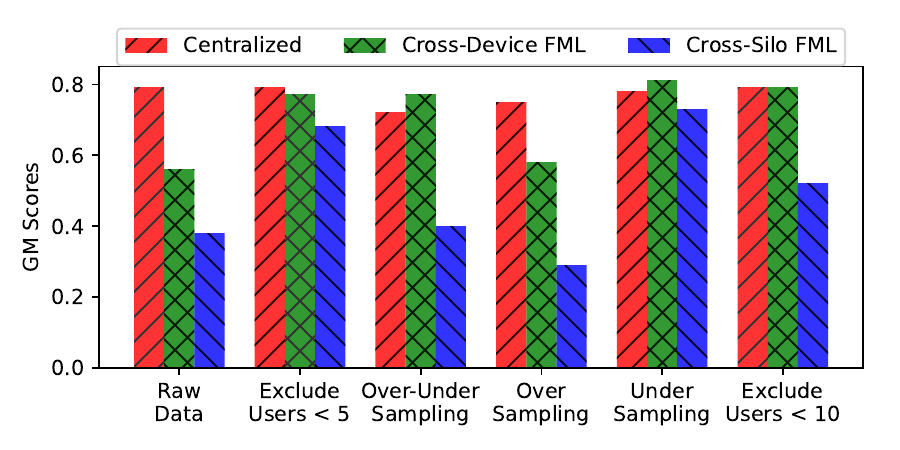}}
\caption{Comparison of the obtained Geometric Mean scores for different learning settings under various experimental scenarios.}
\label{fig:settings_comparison}
\end{figure}

To summarize, our experiments suggest that centralized learning consistently provides robust predictions, regardless of the applied data selection and class imbalance strategies. Nevertheless, cross-device FML presents comparable predictive accuracy to centralized learning. Notably, when under sampling is integrating into cross-device FML, the federated setting surpass the predictive performance of the centralized approach.

Considering the sensitive nature of health and social care data, it is important to consider privacy-enhancing learning approaches such as cross-device FML. By incorporating intelligent client selection mechanisms, the predictive performance of FML can exceed that of centralized approaches. Based on our observations, under sampling is a highly effective technique for improving the performance of FML models.

The comparison of the three different learning settings under various experiment scenarios is shown in Fig.\ref{fig:settings_comparison}. In particular, when we exclude individuals with low information content from the training set, we observe much better results for cross-device FML compared to cross-silo. Furthermore, it should be noted that the results can be further enhanced when the over-under and under sampling techniques are fine-tuned. On the other hand, the use of over sampling techniques leads to a decrease in performance. This behavior aligns consistently with the findings from the cross-device FML experiment.

These observations highlight the importance of individualized approaches in FML. By focusing on users with data of high quality and optimizing data or user selection techniques, FML-based training can surpass centralized approaches. The contrasting results obtained from different re-sampling techniques further emphasize the significance of careful selection of preprocessing strategies to align with the distributed learning paradigm.

\section{Conclusions and Future Work}
\label{conclusion}
In this paper, we have explored the effectiveness of Federated Machine Learning (FML) when applied to a real-world scenario in the health and social care domain. We compared two distributed learning settings, namely cross-device and cross-silo FML, to traditional centralized learning. Our empirical findings suggest that the careful data selection or the utilization of re-sampling techniques can significantly improve the predictive performance of federated learning methods. On the other hand, such techniques did not affect the accuracy of centralized learning. Furthermore, our analysis shows that in specific scenarios, particularly when using under sampling within the cross-device federated setting, FML-based training can surpass other methods in terms of predictive accuracy. This finding indicates that federated learning not only minimizes privacy concerns but with the application of intelligent data selection or re-sampling algorithms, it can outperform centralized approaches, thereby achieving highly accurate predictions in the context of predicting early user dropouts from healthy ageing applications.

Future research could delve deeper into understanding the optimal threshold for excluding users from training, identify the most appropriate re-sampling method for each use case and explore alternative client selection techniques to enhance the model's predictive performance. By gaining a better understanding of the underlying factors that influence the predictive accuracy in federated settings, we can refine preprocessing strategies and improve the overall efficacy of machine learning models applied in real-world scenarios.



\bibliographystyle{IEEEtran}
\bibliography{Bibliography}

\begin{thebibliography}{10}
\providecommand{\url}[1]{#1}
\csname url@samestyle\endcsname
\providecommand{\newblock}{\relax}
\providecommand{\bibinfo}[2]{#2}
\providecommand{\BIBentrySTDinterwordspacing}{\spaceskip=0pt\relax}
\providecommand{\BIBentryALTinterwordstretchfactor}{4}
\providecommand{\BIBentryALTinterwordspacing}{\spaceskip=\fontdimen2\font plus
\BIBentryALTinterwordstretchfactor\fontdimen3\font minus
  \fontdimen4\font\relax}
\providecommand{\BIBforeignlanguage}[2]{{%
\expandafter\ifx\csname l@#1\endcsname\relax
\typeout{** WARNING: IEEEtran.bst: No hyphenation pattern has been}%
\typeout{** loaded for the language `#1'. Using the pattern for}%
\typeout{** the default language instead.}%
\else
\language=\csname l@#1\endcsname
\fi
#2}}
\providecommand{\BIBdecl}{\relax}
\BIBdecl

\bibitem{colizzi_prevention_2020}
\BIBentryALTinterwordspacing
M.~Colizzi, A.~Lasalvia, and M.~Ruggeri, ``Prevention and early intervention in
  youth mental health: is it time for a multidisciplinary and trans-diagnostic
  model for care?'' \emph{International Journal of Mental Health Systems},
  vol.~14, no.~1, p.~23, Mar. 2020. [Online]. Available:
  \url{https://doi.org/10.1186/s13033-020-00356-9}
\BIBentrySTDinterwordspacing

\bibitem{TRIANTAFYLLIDIS2015743}
\BIBentryALTinterwordspacing
A.~Triantafyllidis, C.~Velardo, T.~Chantler, S.~A. Shah, and C.~Paton, ``A
  personalised mobile-based home monitoring system for heart failure: The
  support-hf study,'' \emph{International Journal of Medical Informatics},
  vol.~84, no.~10, pp. 743--753, 2015. [Online]. Available:
  \url{https://www.sciencedirect.com/science/article/pii/S1386505615000969}
\BIBentrySTDinterwordspacing

\bibitem{wiens_no_2019}
\BIBentryALTinterwordspacing
J.~Wiens, S.~Saria, M.~Sendak, M.~Ghassemi, and V.~X. Liu,
  ``\BIBforeignlanguage{en}{Do no harm: a roadmap for responsible machine
  learning for health care},'' \emph{\BIBforeignlanguage{en}{Nature Medicine}},
  vol.~25, no.~9, pp. 1337--1340, Sep. 2019. [Online]. Available:
  \url{https://www.nature.com/articles/s41591-019-0548-6}
\BIBentrySTDinterwordspacing

\bibitem{burns_harnessing_2011}
\BIBentryALTinterwordspacing
M.~N. Burns, M.~Begale, J.~Duffecy, D.~Gergle, and C.~J. Karr,
  ``\BIBforeignlanguage{EN}{Harnessing {Context} {Sensing} to {Develop} a
  {Mobile} {Intervention} for {Depression}},''
  \emph{\BIBforeignlanguage{EN}{Journal of Medical Internet Research}},
  vol.~13, no.~3, p. e1838, Aug. 2011. [Online]. Available:
  \url{https://www.jmir.org/2011/3/e55}
\BIBentrySTDinterwordspacing

\bibitem{QIN2020104121}
\BIBentryALTinterwordspacing
F.-Y. Qin, Z.-Q. Lv, D.-N. Wang, B.~Hu, and C.~Wu, ``Health status prediction
  for the elderly based on machine learning,'' \emph{Archives of Gerontology
  and Geriatrics}, vol.~90, p. 104121, 2020. [Online]. Available:
  \url{https://www.sciencedirect.com/science/article/pii/S0167494320301151}
\BIBentrySTDinterwordspacing

\bibitem{Voigt_2017}
\BIBentryALTinterwordspacing
P.~Voigt and A.~von~dem Bussche, \emph{The {EU} General Data Protection
  Regulation ({GDPR})}.\hskip 1em plus 0.5em minus 0.4em\relax Springer
  International Publishing, 2017. [Online]. Available:
  \url{https://doi.org/10.1007/978-3-319-57959-7}
\BIBentrySTDinterwordspacing

\bibitem{pmlr-v54-mcmahan17a}
\BIBentryALTinterwordspacing
B.~McMahan, E.~Moore, D.~Ramage, S.~Hampson, and B.~A.~y. Arcas,
  ``{Communication-Efficient Learning of Deep Networks from Decentralized
  Data},'' in \emph{Proceedings of the 20th International Conference on
  Artificial Intelligence and Statistics}, ser. Proceedings of Machine Learning
  Research, A.~Singh and J.~Zhu, Eds., vol.~54.\hskip 1em plus 0.5em minus
  0.4em\relax PMLR, 20--22 Apr 2017, pp. 1273--1282. [Online]. Available:
  \url{https://proceedings.mlr.press/v54/mcmahan17a.html}
\BIBentrySTDinterwordspacing

\bibitem{10.1145/3133956.3133982}
\BIBentryALTinterwordspacing
K.~Bonawitz, V.~Ivanov, B.~Kreuter, A.~Marcedone, H.~B. McMahan, S.~Patel,
  D.~Ramage, A.~Segal, and K.~Seth, ``Practical secure aggregation for
  privacy-preserving machine learning,'' in \emph{Proceedings of the 2017 ACM
  SIGSAC Conference on Computer and Communications Security}, ser. CCS
  '17.\hskip 1em plus 0.5em minus 0.4em\relax New York, NY, USA: Association
  for Computing Machinery, 2017, p. 1175–1191. [Online]. Available:
  \url{https://doi.org/10.1145/3133956.3133982}
\BIBentrySTDinterwordspacing

\bibitem{ZHU2021371}
\BIBentryALTinterwordspacing
H.~Zhu, J.~Xu, S.~Liu, and Y.~Jin, ``Federated learning on non-iid data: A
  survey,'' \emph{Neurocomputing}, vol. 465, pp. 371--390, 2021. [Online].
  Available:
  \url{https://www.sciencedirect.com/science/article/pii/S0925231221013254}
\BIBentrySTDinterwordspacing

\bibitem{9835537}
Q.~Li, Y.~Diao, Q.~Chen, and B.~He, ``Federated learning on non-iid data silos:
  An experimental study,'' in \emph{2022 IEEE 38th International Conference on
  Data Engineering (ICDE)}, 2022, pp. 965--978.

\bibitem{perifanis2022federated}
\BIBentryALTinterwordspacing
V.~Perifanis, N.~Pavlidis, R.-A. Koutsiamanis, and P.~S. Efraimidis,
  ``Federated learning for 5g base station traffic forecasting,''
  \emph{Computer Networks}, vol. 235, p. 109950, 2023. [Online]. Available:
  \url{https://www.sciencedirect.com/science/article/pii/S138912862300395X}
\BIBentrySTDinterwordspacing

\bibitem{pmlr-v151-jee-cho22a}
Y.~Jee~Cho, J.~Wang, and G.~Joshi, ``Towards understanding biased client
  selection in federated learning,'' in \emph{Proceedings of The 25th
  International Conference on Artificial Intelligence and Statistics}, ser.
  Proceedings of Machine Learning Research, G.~Camps-Valls, F.~J.~R. Ruiz, and
  I.~Valera, Eds., vol. 151.\hskip 1em plus 0.5em minus 0.4em\relax PMLR,
  28--30 Mar 2022, pp. 10\,351--10\,375.

\bibitem{PRATI2022104791}
\BIBentryALTinterwordspacing
G.~Prati, ``Correlates of quality of life, happiness and life satisfaction
  among european adults older than 50 years: A machine‐learning approach,''
  \emph{Archives of Gerontology and Geriatrics}, vol. 103, p. 104791, 2022.
  [Online]. Available:
  \url{https://www.sciencedirect.com/science/article/pii/S0167494322001789}
\BIBentrySTDinterwordspacing

\bibitem{torous_smartphones_2018}
\BIBentryALTinterwordspacing
J.~Torous, M.~E. Larsen, C.~Depp, T.~D. Cosco, I.~Barnett, M.~K. Nock, and
  J.~Firth, ``\BIBforeignlanguage{en}{Smartphones, {Sensors}, and {Machine}
  {Learning} to {Advance} {Real}-{Time} {Prediction} and {Interventions} for
  {Suicide} {Prevention}: a {Review} of {Current} {Progress} and {Next}
  {Steps}},'' \emph{\BIBforeignlanguage{en}{Current Psychiatry Reports}},
  vol.~20, no.~7, p.~51, Jun. 2018. [Online]. Available:
  \url{https://doi.org/10.1007/s11920-018-0914-y}
\BIBentrySTDinterwordspacing

\bibitem{okeeffe_predicting_2018}
S.~O'Keeffe, P.~Martin, I.~M. Goodyer, P.~Wilkinson, I.~Consortium, and
  N.~Midgley, ``\BIBforeignlanguage{eng}{Predicting dropout in adolescents
  receiving therapy for depression},''
  \emph{\BIBforeignlanguage{eng}{Psychotherapy Research: Journal of the Society
  for Psychotherapy Research}}, vol.~28, no.~5, pp. 708--721, Sep. 2018.

\bibitem{Fico_2023_MAHA}
G.~Fico, P.~Abril-Jimenez, I.~Lombroni, B.~Merino-Barbancho, B.~Patricio,
  P.~Arroyo, G.~Cea, G.~Mejias-Izquierdo, A.~Medrano, M.~F. Cabrera-Umpierrez,
  J.~Henriques, P.~Carvalho, A.~Mata, and M.~T. Arredondo~Waldmeyer, ``The maha
  dataset: Understanding and improving adherence to digital interventions for
  active and healthy ageing,'' in \emph{World Congress on Medical Physics and
  Biomedical Engineering 2022}.\hskip 1em plus 0.5em minus 0.4em\relax Springer
  Singapore, 2023.

\bibitem{rieke_future_2020}
\BIBentryALTinterwordspacing
N.~Rieke, J.~Hancox, W.~Li, F.~Milletarì, H.~R. Roth, S.~Albarqouni, S.~Bakas,
  M.~N. Galtier, B.~A. Landman, K.~Maier-Hein, S.~Ourselin, M.~Sheller, R.~M.
  Summers, A.~Trask, D.~Xu, M.~Baust, and M.~J. Cardoso,
  ``\BIBforeignlanguage{en}{The future of digital health with federated
  learning},'' \emph{\BIBforeignlanguage{en}{npj Digital Medicine}}, vol.~3,
  no.~1, pp. 1--7, Sep. 2020. [Online]. Available:
  \url{https://www.nature.com/articles/s41746-020-00323-1}
\BIBentrySTDinterwordspacing

\bibitem{BRISIMI201859}
\BIBentryALTinterwordspacing
T.~S. Brisimi, R.~Chen, T.~Mela, A.~Olshevsky, I.~C. Paschalidis, and W.~Shi,
  ``Federated learning of predictive models from federated electronic health
  records,'' \emph{International Journal of Medical Informatics}, vol. 112, pp.
  59--67, 2018. [Online]. Available:
  \url{https://www.sciencedirect.com/science/article/pii/S138650561830008X}
\BIBentrySTDinterwordspacing

\bibitem{dayan_federated_2021}
\BIBentryALTinterwordspacing
I.~Dayan, H.~R. Roth, A.~Zhong, A.~Harouni, A.~Gentili, A.~Z. Abidin, and
  A.~Liu, ``\BIBforeignlanguage{en}{Federated learning for predicting clinical
  outcomes in patients with {COVID}-19},'' \emph{\BIBforeignlanguage{en}{Nature
  Medicine}}, vol.~27, no.~10, pp. 1735--1743, Oct. 2021. [Online]. Available:
  \url{https://www.nature.com/articles/s41591-021-01506-3}
\BIBentrySTDinterwordspacing

\bibitem{adnan_federated_2022}
\BIBentryALTinterwordspacing
M.~Adnan, S.~Kalra, J.~C. Cresswell, G.~W. Taylor, and H.~R. Tizhoosh,
  ``\BIBforeignlanguage{en}{Federated learning and differential privacy for
  medical image analysis},'' \emph{\BIBforeignlanguage{en}{Scientific
  Reports}}, vol.~12, no.~1, p. 1953, Feb. 2022. [Online]. Available:
  \url{https://www.nature.com/articles/s41598-022-05539-7}
\BIBentrySTDinterwordspacing

\bibitem{9900733}
I.~Diamantoulaki, P.~D. Diamantoulakis, P.~S. Bouzinis, P.~Sarigiannidis, and
  G.~K. Karagiannidis, ``Health risk assessment with federated learning,'' in
  \emph{2022 International Balkan Conference on Communications and Networking
  (BalkanCom)}, 2022, pp. 57--61.

\bibitem{Perifanis_2023_Dropout}
V.~Perifanis, I.~Michailidi, G.~Stamatelatos, G.~Drosatos, and P.~S.
  Efraimidis, ``Predicting early dropouts of an active and healthy ageing
  app,'' in \emph{World Congress on Medical Physics and Biomedical Engineering
  2022}.\hskip 1em plus 0.5em minus 0.4em\relax Springer Singapore, 2023.

\bibitem{CAI201870}
\BIBentryALTinterwordspacing
J.~Cai, J.~Luo, S.~Wang, and S.~Yang, ``Feature selection in machine learning:
  A new perspective,'' \emph{Neurocomputing}, vol. 300, pp. 70--79, 2018.
  [Online]. Available:
  \url{https://www.sciencedirect.com/science/article/pii/S0925231218302911}
\BIBentrySTDinterwordspacing

\bibitem{DOUZAS20181}
\BIBentryALTinterwordspacing
G.~Douzas, F.~Bacao, and F.~Last, ``Improving imbalanced learning through a
  heuristic oversampling method based on k-means and smote,'' \emph{Information
  Sciences}, vol. 465, pp. 1--20, 2018. [Online]. Available:
  \url{https://www.sciencedirect.com/science/article/pii/S0020025518304997}
\BIBentrySTDinterwordspacing

\bibitem{10.1145/1007730.1007735}
\BIBentryALTinterwordspacing
G.~E. A. P.~A. Batista, R.~C. Prati, and M.~C. Monard, ``A study of the
  behavior of several methods for balancing machine learning training data,''
  \emph{SIGKDD Explor. Newsl.}, vol.~6, no.~1, p. 20–29, jun 2004. [Online].
  Available: \url{https://doi.org/10.1145/1007730.1007735}
\BIBentrySTDinterwordspacing

\bibitem{9306745}
G.~Drainakis, K.~V. Katsaros, P.~Pantazopoulos, V.~Sourlas, and A.~Amditis,
  ``Federated vs. centralized machine learning under privacy-elastic users: A
  comparative analysis,'' in \emph{2020 IEEE 19th International Symposium on
  Network Computing and Applications (NCA)}, 2020, pp. 1--8.

\bibitem{srivastava2014dropout}
N.~Srivastava, G.~Hinton, A.~Krizhevsky, I.~Sutskever, and R.~Salakhutdinov,
  ``Dropout: a simple way to prevent neural networks from overfitting,''
  \emph{The journal of machine learning research}, vol.~15, no.~1, pp.
  1929--1958, 2014.

\bibitem{beutel2020flower}
D.~J. Beutel, T.~Topal, A.~Mathur, X.~Qiu, J.~Fernandez-Marques, Y.~Gao,
  L.~Sani, K.~H. Li, T.~Parcollet, P.~P.~B. de~Gusm{\~a}o \emph{et~al.},
  ``Flower: A friendly federated learning research framework,'' \emph{arXiv
  preprint arXiv:2007.14390}, 2020.

\bibitem{tensorflow2015-whitepaper}
\BIBentryALTinterwordspacing
M.~Abadi, A.~Agarwal, P.~Barham, E.~Brevdo, Z.~Chen, C.~Citro, and G.~S.
  Corrado, ``{TensorFlow}: Large-scale machine learning on heterogeneous
  systems,'' 2015, software available from tensorflow.org. [Online]. Available:
  \url{https://www.tensorflow.org/}
\BIBentrySTDinterwordspacing

\end{thebibliography}


\end{document}